\documentclass{article}

 \usepackage[preprint]{nips_2018}

\usepackage[utf8]{inputenc} 
\usepackage[T1]{fontenc}    
\usepackage{hyperref}       
\usepackage{url}            
\usepackage{booktabs}       
\usepackage{amsfonts}       
\usepackage{nicefrac}       
\usepackage{microtype}      
\usepackage{bbm}
\usepackage{graphicx}
\usepackage{caption}
\usepackage{subcaption}
\usepackage[ruled]{algorithm2e}
\usepackage{algpseudocode}

\newcommand{\identityFunction}{ \mathbbm{1}}
\newcommand{\expected}{ \mathbbm{E}}
\title{Towards a Simple Approach to\\ Multi-step Model-based Reinforcement Learning}

\author{
  Kavosh Asadi \\
  Department of Computer Science\\
  Brown University\\
\And
  Evan Cater \\
  Department of Computer Science\\
  Brown University\\
\And
  Dipendra Misra \\
  Department of Computer Science\\
  Cornell University\\
\And
  Michael L. Littman \\
  Department of Computer Science\\
  Brown University\\
}

\begin{document}

\maketitle

\begin{abstract}
When environmental interaction is expensive, model-based reinforcement learning offers a solution by planning ahead and avoiding costly mistakes. Model-based agents typically learn a single-step transition model. In this paper, we propose a multi-step model that predicts the outcome of an action sequence with variable length. We show that this model is easy to learn, and that the model can make policy-conditional predictions. We report preliminary results that show a clear advantage for the multi-step model compared to its one-step counterpart.
\end{abstract}
\section{Introduction}
Reinforcement learning \citep{Sutton_the_book} has become the framework for studying AI agents that learn, plan, and act in uncertain and sequential environments. This success is partly due to simple and general reinforcement learning algorithms that, when equipped with neural networks, can solve complex domains such as games with massive state spaces \citep{Mnih_Atari,Silver_Go}. Games usually provide a simulator which can be used to experience a large amount of environmental interaction. A good simulator is unavailable in many applications, so it is imperative to learn a good behavior with fewer environmental interactions.

When environmental interaction is expensive but computation is cheap, the model-based approach to reinforcement learning offers a solution by summarizing past experience using a \textit{model} informally thought of as an internal simulator of the environment \citep{Sutton_the_book}. An agent can then use the learned model in different capacities. For example, a model can be used to perform tree search \citep{UCT,Silver_Go}, to update the value-function estimate \citep{Dyna_original,parr_linear_models,linear_dyna}, to update an explicitly represented policy \citep{abbeel_inaccurate_models,Pilco}, or to make option-conditional predictions \citep{options,liner_options}. 

Regardless of how a model is utilized, effectiveness of planning depends on the accuracy of the model. A widely accepted view is that all models are imperfect, but some are still useful \citep{all_models_are_wrong}. Generally speaking, modeling errors can be due to overfitting or underfitting, or simply due to the fact that the setting is agnostic, and so, some irreducible error exists. In reinforcement learning, for example, partial observability and non-Markovian dynamics can yield an agnostic setting.

In model-based setting it is common to learn a one-step model of the environment. When performing rollouts using a one-step model, the output of the model is then used in the subsequent step as the input. It is discovered, however, that in this case even small modeling errors, which are unavoidable as argued above, can severly degrade multi-step predictions \citep{Talvitie_self_correcting,time_series_error_compound,Asadi_Lipschitz}. Intuitively, the problem is that the model is not trained to perform reasonably on its own output. The rollout process can then get derailed by moving out of the state space after a few steps.

Here we study a direct approach to computing multi-step predictions. We propose a simple multi-step model that learns to predict the outcome of an action sequence with variable length. We show that in terms of prediction accuracy our proposed model outperforms a one-step model when the prediction horizon is large. We further show that the model can provide the outcome of running a policy rather than just a fixed sequence of actions. To show the effectiveness of the multi-step model, we use it for value-function optimization in the context of actor-critic reinforcement learning. Moreover, we report preliminary results on Atari Breakout showing that the multi-step model outperforms the one-step model in terms of frame prediction after multiple timesteps.
\section{Background and Notation}
We briefly present the background and the notation used to articulate our results. For a complete background on reinforcement learning see \cite{Sutton_the_book}, and for a complete background on MDPs see \cite{puterman_mdp}.

We focus on the reinforcement learning problem in which an agent interacts with an environment to maximize long-term reward. The problem is formulated by a Markov Decision Process, or simply an MDP. The tuple $\prec \mathcal{S},\mathcal{A},R,T,\gamma\succ$ represents an MDP where we assume a continuous state-space $\mathcal{S}$, and a discrete action-space $\mathcal{A}$. Our goal is to find a policy $\pi:\mathcal{S}\mapsto\textrm{Pr}(\mathcal{A})$ that achieves high $\gamma$-discounted cumulative reward.

Here $R$ and $T$ represent the reward and transition dynamics of the MDP. More specifically $R:\mathcal{S}\times\mathcal{A}\times\mathcal{S} \mapsto \mathbb{R}$ is the reward after taking an action $a\in\mathcal{A}$ in a state $s\in\mathcal{S}$ and moving to a next state $s'\in\mathcal{S}$. This is denoted by $R(s,a,s')$. For transition dynamics we use an overloaded notation. In the simplest case, transition function can take as input a state, a next state, and an action:
$$T^{1}(s,s',a):=\textrm{Pr}(s_{t+1}=s'|s_{t}=s,a_{t}=a)\ .$$
Similarly, we define a transition function conditioned on a state and a sequence of actions:
$$T^{n}(s,s',a_{0},a_{1},\cdots,  a_{n-1}):=\textrm{Pr}(s_{t+n}=s'|s_{t}=s,a_{t}=a_{0},a_{t+1}=a_{1},...,a_{t+n-1}=a_{n-1})\ ,$$
as well as conditioned on a state and a policy:
$$T^{n}(s,s',\pi):=\textrm{Pr}(s_{t+n}=s'|s_{t}=s,\pi)\ .$$
It is useful to compute the long-term goodness of taking an action $a$ in a state $s$. Referred to as the value function, this is defined as follows:
$$Q^{\pi}(s,a):=\expected \big[\sum_{i=0}^{T}\gamma^{i} R(s_{i},a_{i},s_{i+1})|s_{0}=s,a_{0}=a,\pi\big] \ .$$
\section{Policy-Conditional Prediction}
In this section we study the following basic question: How can we learn to predict the outcome of executing a policy $\pi$ in a state $s$ given a horizon $n$. More specifically, our aim is to compute (or sample from) the following distribution:
$$T^{n}(s,s',\pi):=\textrm{Pr}(s_{t+n}=s'|s_{t}=s,\pi) \ .$$
We explain the standard approach in the next section, and introduce our approach in the subsequent section.
\subsection{Single-step Model}
The common way of making the $n$-step prediction is to break the problem into $n$ single-step prediction problems as follows:
\begin{eqnarray*}
T^{n}(s,s',\pi)&:=&\textrm{Pr}(s_{t+n}=s'|s_{t}=s,\pi)\\
&=&\sum_{s_{t+1}} ... \sum_{s_{t+n-1}}\textrm{Pr}(s_{t+n}=s',s_{t+n-1},...,s_{t+1}|s_{t}=s,\pi)\\
&=&\sum_{s_{t+1}} ... \sum_{s_{t+n-1}}\textrm{Pr}(s_{t+n}=s'|s_{t+n-1},...,s_{t}=s,\pi)\ ...\ \textrm{Pr}(s_{t+1}|s_{t}=s,\pi)\\
&=&\sum_{s_{t+1}}\textrm{Pr}(s_{t+1}|s_{t}=s,\pi)\ ... \sum_{s_{t+n-1}}\textrm{Pr}(s_{t+n}=s'|s_{t+n-1},\pi)\quad \textrm{(Markovian property)}\\
&=&\sum_{s_{t+1}}T^{1}(s,s_{t+1},\pi)\ ... \sum_{s_{t+n-1}}T^{1}(s_{t+n-1},s',\pi)\ .\\
\end{eqnarray*}
The one-step probability distribution could further be written as:
$$T^{1}(s,s',\pi):=\sum_{a}\pi(a|s)\ T^{1}(s,s',a)\ .$$
Note that in a continuous state-space it is easier to work with a deterministic model, rather than the full distribution, so following prior work \citep{parr_linear_models,linear_dyna} we use a deterministic approximation that learns to output the expected next state by minimizing mean squared error:
$$\widehat{T}^{1}(s,a)\approx \expected [s_{t+1}|s_{t}=s,a_{t}=a] \ .$$
We represent this model, as well as all parameterized functions, using deep neural networks \cite{deep_learning_nature}. This approximation can be poor in general, but is theoretically justified even when environment is stochastic \citep{parr_linear_models,linear_dyna}.

\begin{figure}
    \centering
    \includegraphics[width=0.7\textwidth]{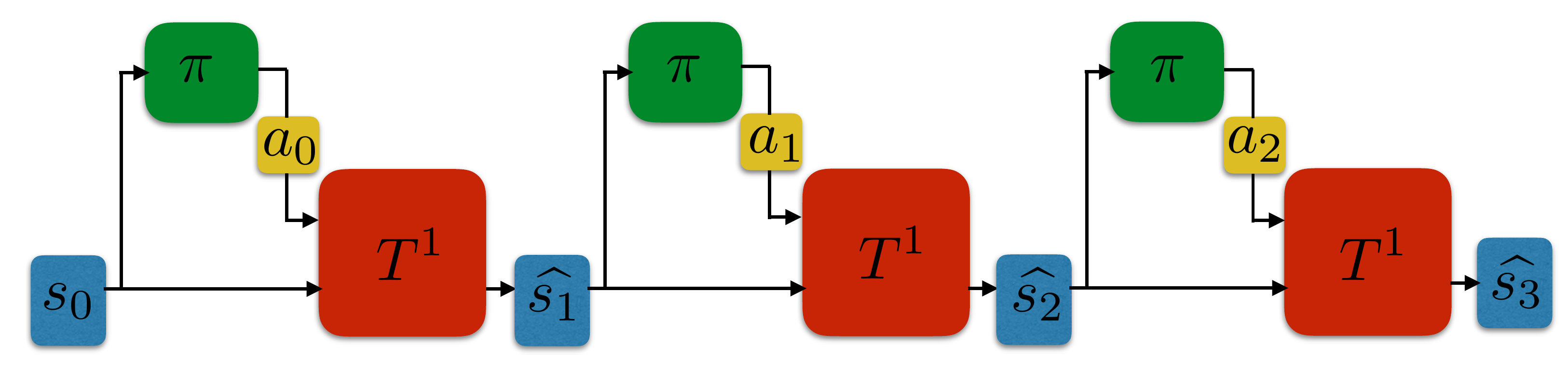}
    \caption{A 3-step rollout using a one-step model.}
    \label{fig:roll_out_one_step}
\end{figure}

Finally, using this model, we can easily sample a state $n$ steps into the future:
$$\widehat s_n = \widehat{T}^1(\ ...\ \widehat{T}^1(\widehat{T}^1(s_0,a_0),a_1),...,a_{n-1})\quad \textrm{where}\quad a_i\sim \pi\big(\cdot|\widehat{T}^1(\ ...\ \widehat{T}^1(\widehat{T}^1(s_0,a_0),a_1),...,a_{i-1})\big)$$
Figure \ref{fig:roll_out_one_step} illustrates the rollout process. The downside of this approach is that model's output is provided as input $n-1$ times, yet the model is not trained on its own output. In practice, this causes the compounding error problem.
\begin{figure}
    \centering
    \includegraphics[width=0.7\textwidth]{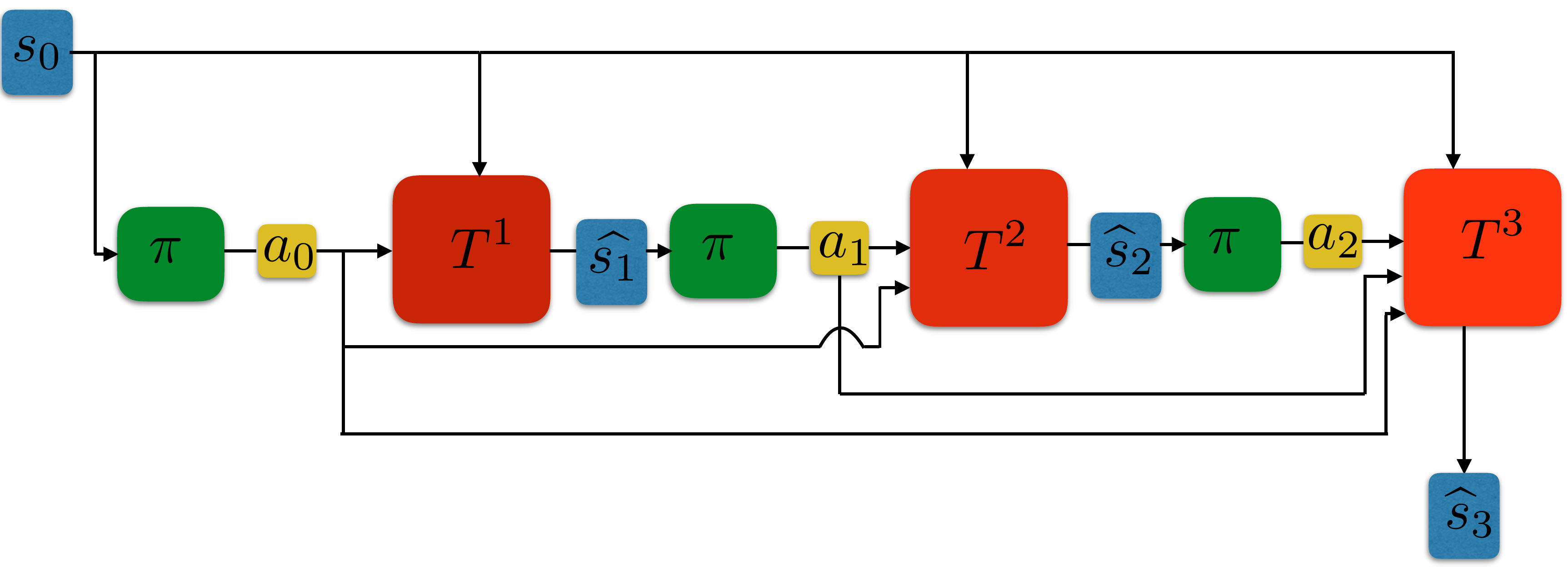}
    \caption{A 3-step rollout using a multi-step model.}
    \label{fig:roll_out_multi_step}
\end{figure}
\subsection{Multi-step Model}
We now articulate an alternative approach to computing multi-step predictions. As we will later on show, it is quite easy to use experience to learn models conditioned on action sequences. So key to our approach is to rewrite the policy-conditional prediction $T^{n}(s,s',\pi)$ in terms of predictions conditioned on action sequences.

\begin{eqnarray*}
\!T^{n}(s,s',\pi)&:=&\textrm{Pr}(s_{t+n}=s'|s_{t}=s,\pi)\\
&\approx&\!\sum_{a_t} ...\!\sum_{a_{t+n-1}}\!\textrm{Pr}(a_{t},\ ...\ ,a_{t+n-1}|s_{t}=s,\pi)\ \identityFunction\big(s'=\widehat{T}^{n}(s,a_t,\ ...\ ,a_{t+n-1})\big) \ .
\end{eqnarray*}
This approximation is perfect if the setting is deterministic. Now note that $\widehat{T}^{n}(s,a_t,\ ...\ ,a_{t+n-1})$ is conditioned on an action sequence, so we only need to worry about the other term, namely $\textrm{Pr}(a_{t},...,a_{n+t-1}|s_{t}=s,\pi)$. We can rewrite this as follows:
\begin{eqnarray*}
\textrm{Pr}(a_{t},...,a_{t+n-1}|s_{t}=s,\pi)&=&\textrm{Pr}(a_{t+n-1}|a_{t},...,a_{t+n-2},s_{t}=s,\pi)\textrm{Pr}(a_{t},...,a_{t+n-2}|s_{t}=s,\pi)\\
&=&\textrm{Pr}\big(a_{t+n-1}|T^{n-1}(s,a_{t},...,a_{t+n-2}),\pi\big)\textrm{Pr}(a_{t},...,a_{t+n-2}|s_{t}=s,\pi)\\
&=&\pi\big(a_{t+n-1}|T^{n-1}(s,a_{t},...,a_{t+n-2})\big)\textrm{Pr}(a_{t},...,a_{t+n-2}|s_{t}=s,\pi)\ .\\
\end{eqnarray*}
Continuing for $n-1$ more steps we get:
$$\textrm{Pr}(a_{t},...,a_{t+n-1}|s_{t}=s,\pi)=\pi(a_{t}|s)\prod_{i=1}^{n-1}\pi\big(a_{t+i}|T^{i}(s,a_{t},...,a_{t+i-1})\big)\ .$$
Putting it together, we have:
$$T^{n}(s,s',\pi)=\sum_{a_0}...\sum_{a_{n-1}}\pi(a_{0}|s)\prod_{i=1}^{n-1}\pi\big(a_{t+i}|T^{i}(s,a_{t},...,a_{t+i-1})\big)  \ \identityFunction \big(s'=T^{n}(s,a_t,\ ...\ ,a_{t+n-1})\big) \ .$$
In simple terms, the multi-step model consists of $n$ functions where each function takes as input an action sequence of length $l=1:n$ along with a state, and predicts the state after $l$ steps. Finally, we can easily sample from the model as follows:
$$\widehat s_n = T^{n}(s,a_0,...,a_{n-1})\quad \textrm{where} \quad a_{i}\sim \pi\big(\cdot|T^{i}(s,a_{0},...,a_{i-1})\big)\quad \textrm{with}\quad T^{0}(s):=s $$
We illustrate the rollout process using Figure \ref{fig:roll_out_multi_step}. Note the distinction between the two rollout scenarios. Unlike the one-step case, we never feed the output of a transition function to another transition function, thereby avoiding the compounding error problem.
\section{Experiments}
\begin{figure}
\centering
\begin{minipage}{.65\textwidth}
  \centering
  \includegraphics[width=.9\linewidth]{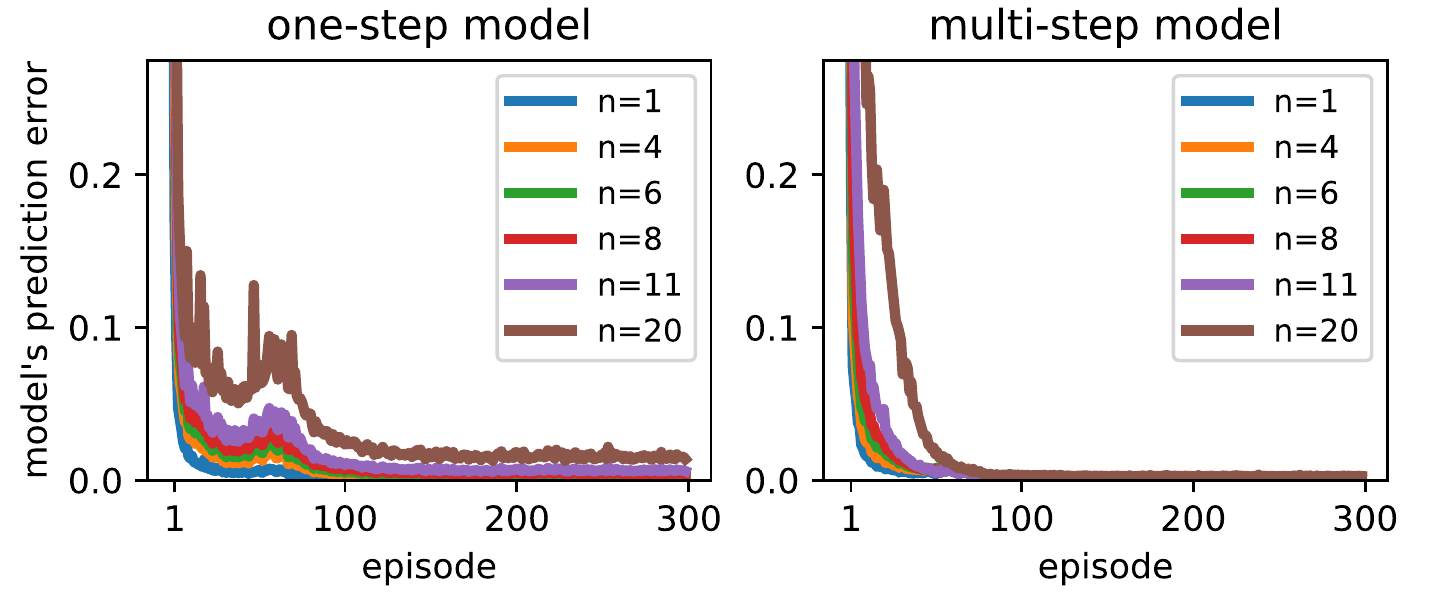}
\end{minipage}%
\begin{minipage}{.35\textwidth}
  \centering
  \includegraphics[width=.9\linewidth]{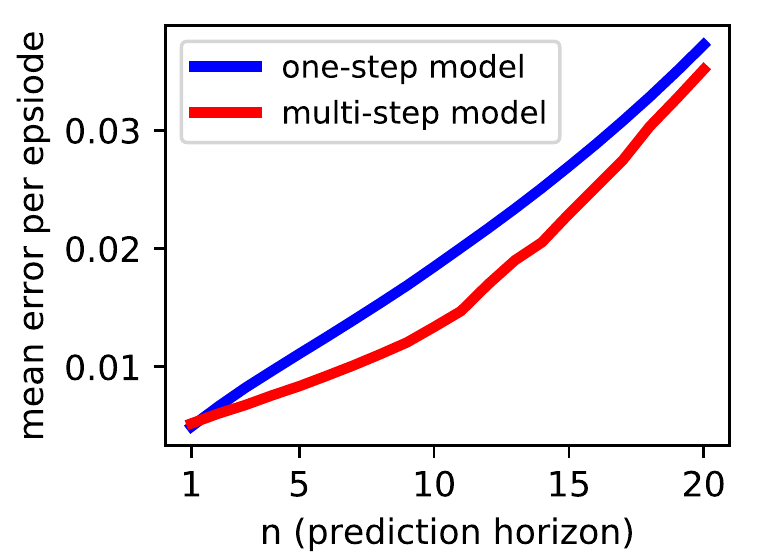}
\end{minipage}
\centering
\begin{minipage}{.65\textwidth}
  \centering
  \includegraphics[width=.9\linewidth]{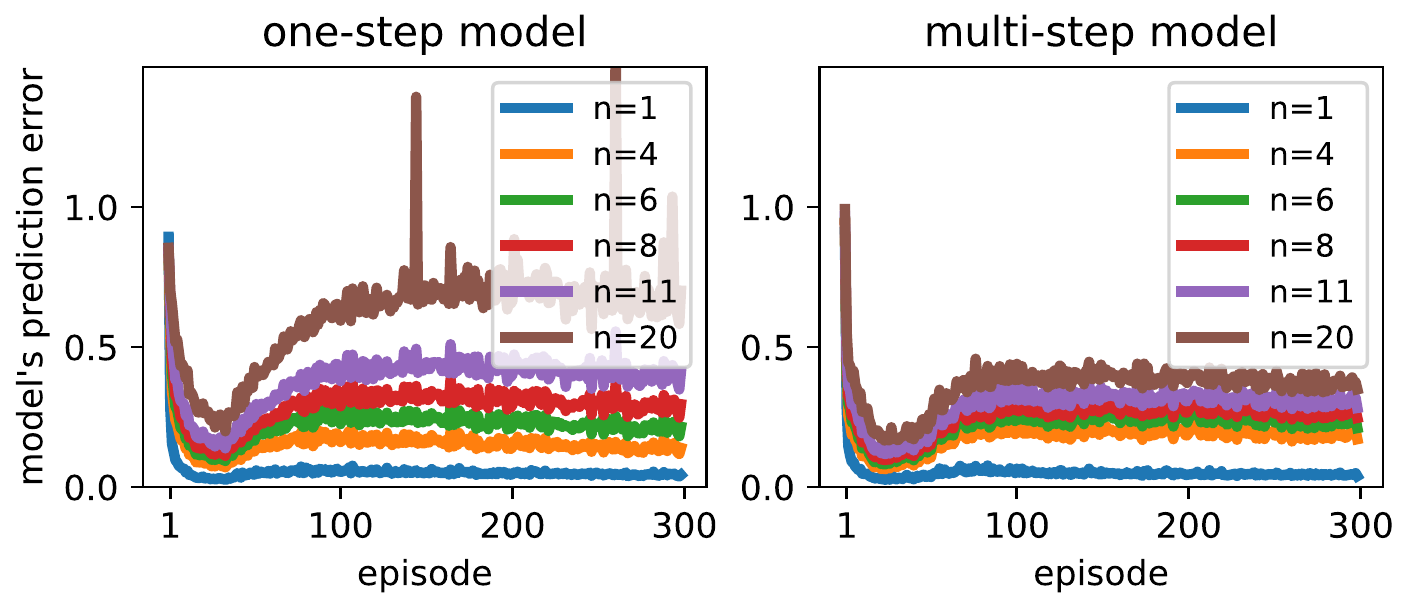}
\end{minipage}%
\begin{minipage}{.35\textwidth}
  \centering
  \includegraphics[width=.9\linewidth]{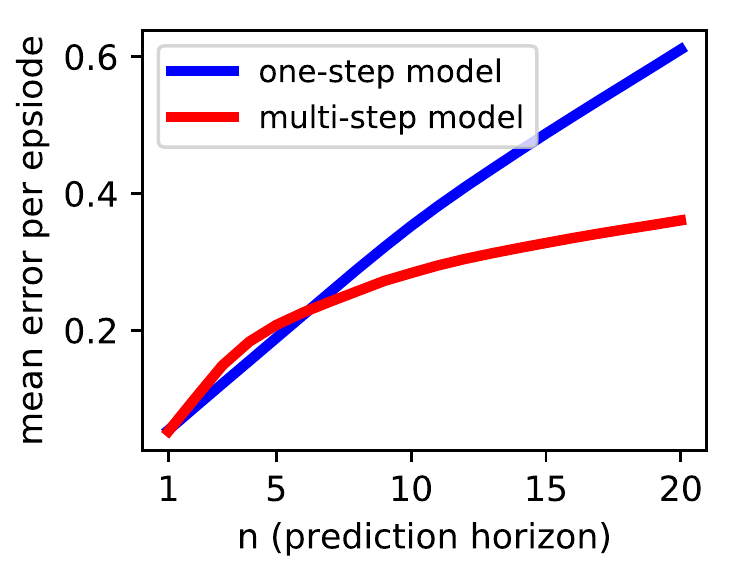}
\end{minipage}
\centering
\begin{minipage}{.65\textwidth}
  \centering
  \includegraphics[width=.9\linewidth]{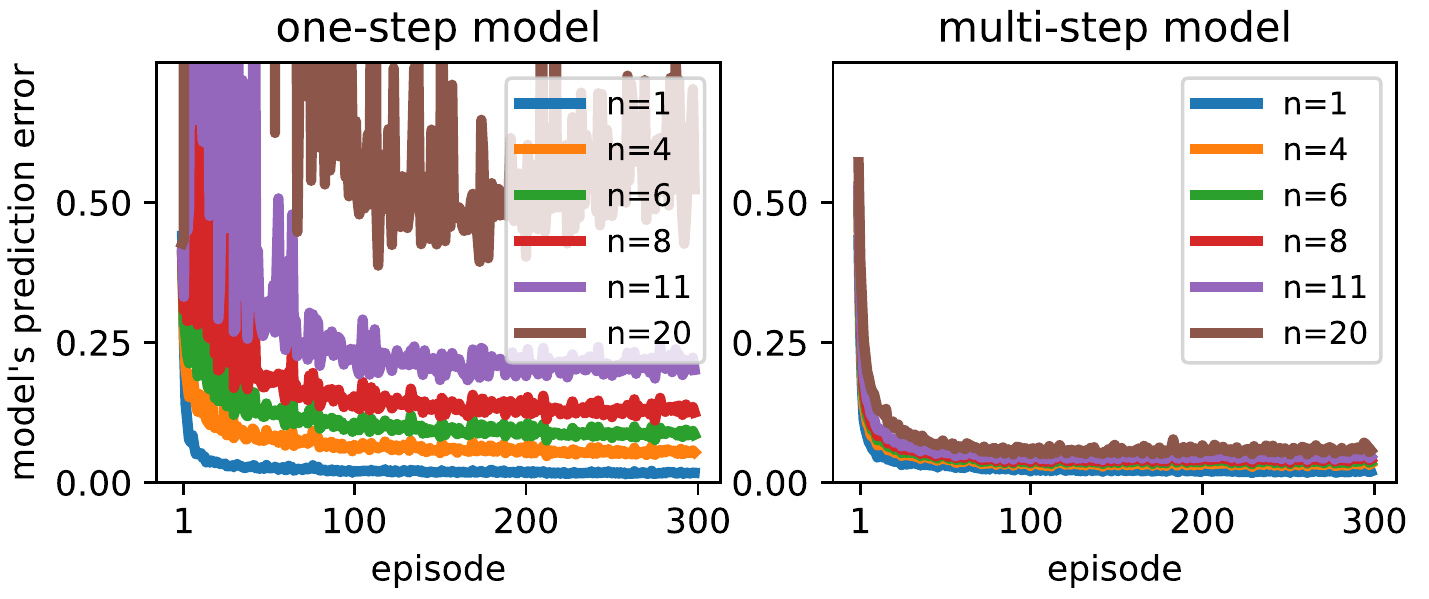}
\end{minipage}%
\begin{minipage}{.35\textwidth}
  \centering
  \includegraphics[width=.9\linewidth]{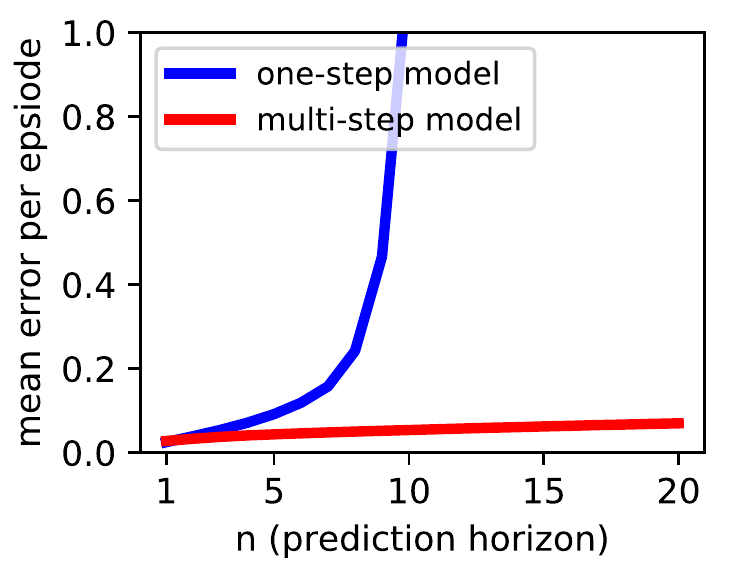}
\end{minipage}
\caption{Accuracy of one-step and multi-step models on Cart Pole (first row), Acrobot (second row), and Lunar Lander (third row). First column shows error when using the one-step model, while the second column shows error when using the multi-step model. Third column is area under the curve for each prediction horizon. Results are averaged over 50 runs. Lower is better.}
\label{model_performance}
\end{figure}
In this section we discuss preliminary empirical results comparing the two models. We evaluate the models in terms of prediction accuracy and usefulness for planning.
\subsection{Control Domains}
We ran our first experiments on three control domains, namely Cart Pole, Acrobot, and Lunar Lander. Public implementation of these domains are available on the web \citep{gym}. We used all-actions version of the actor-critic algorithm \citep{neuronlike_barto,sutton_ac,asadi_mac} as the reinforcement learner where updates to actor and critic were performed at the end of each episode. We report all hyper-parameters in the Appendix.

Our first goal is to compare the accuracy of the two models. To this end, we performed one episode of environmental interaction using the policy network, then measured the accuracy of models on the (just terminated) episode. An episode of experience was used for model training only after model's prediction accuracy was computed for the episode.

In order to measure prediction accuracy for a given horizon $n$ and a given state in the episode, we computed the squared difference between model's prediction and the state observed after $n$ steps. We then took average over all states within an episode. These quantities were separately computed for both models. We present our findings in Figure \ref{model_performance}. The multi-step model generally performed better than the one-step model, other than when the horizon was short in the Acrobot problem.

\begin{figure}
\centering
\begin{minipage}{.65\textwidth}
  \centering
  \includegraphics[width=.9\linewidth]{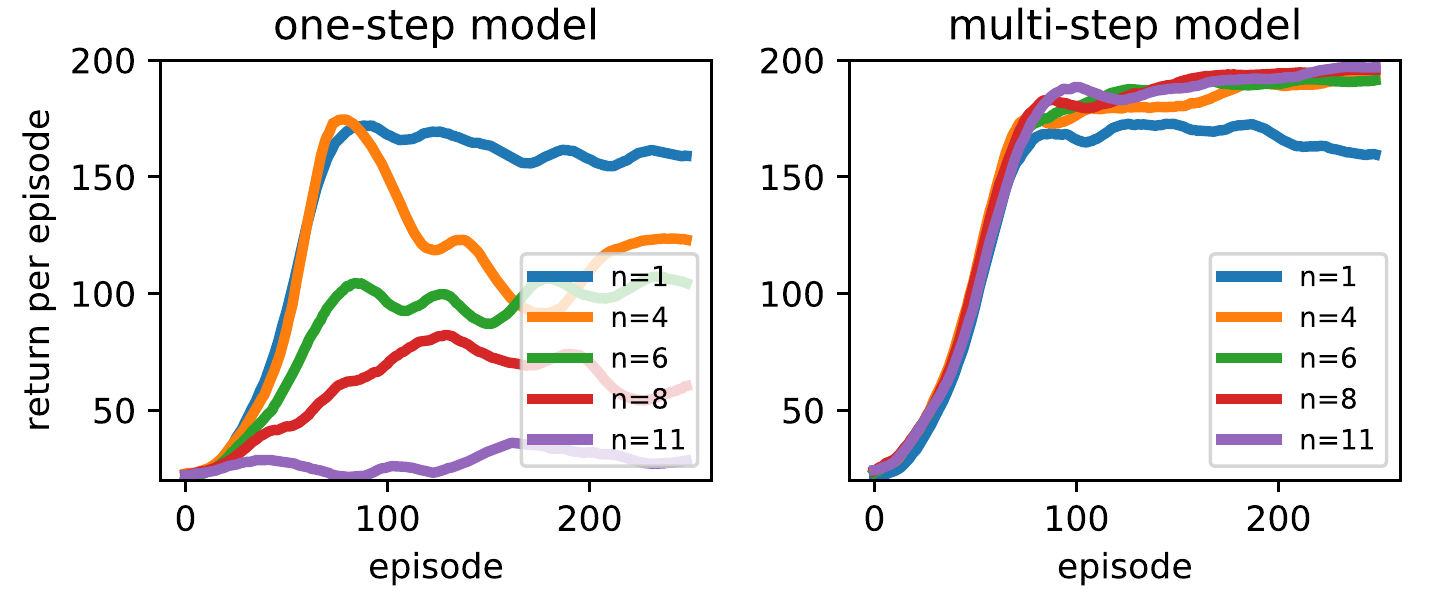}
\end{minipage}%
\begin{minipage}{.35\textwidth}
  \centering
  \includegraphics[width=.9\linewidth]{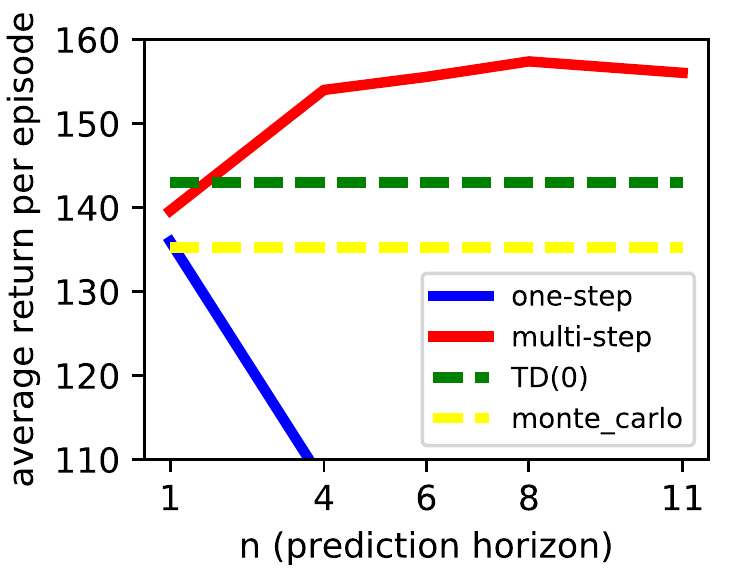}
\end{minipage}
\centering
\begin{minipage}{.65\textwidth}
  \centering
  \includegraphics[width=.9\linewidth]{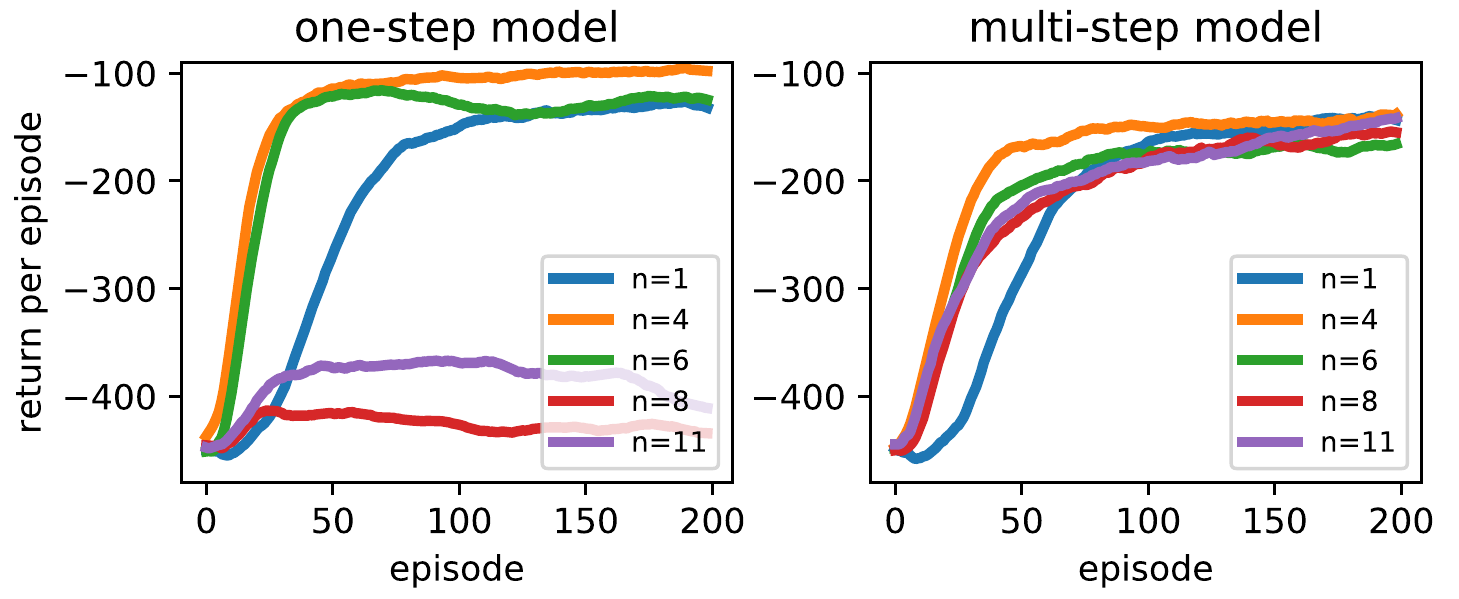}
\end{minipage}%
\begin{minipage}{.35\textwidth}
  \centering
  \includegraphics[width=.9\linewidth]{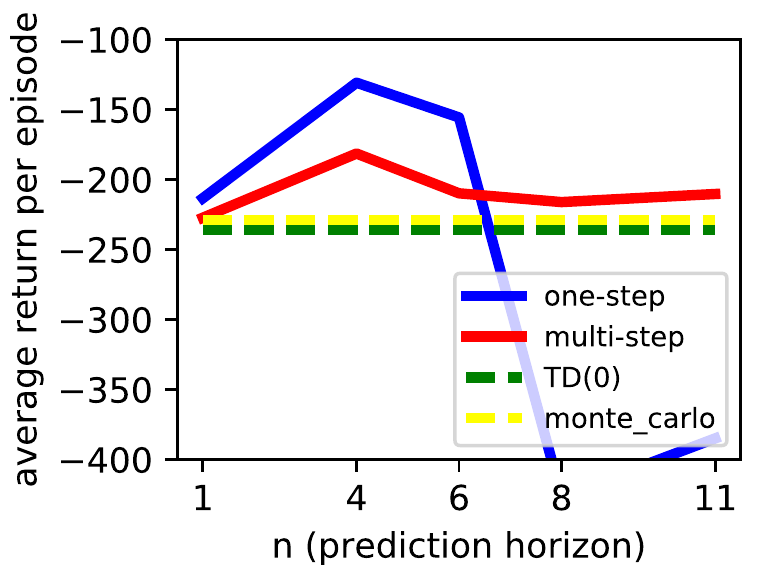}
\end{minipage}
\centering
\begin{minipage}{.65\textwidth}
  \centering
  \includegraphics[width=.9\linewidth]{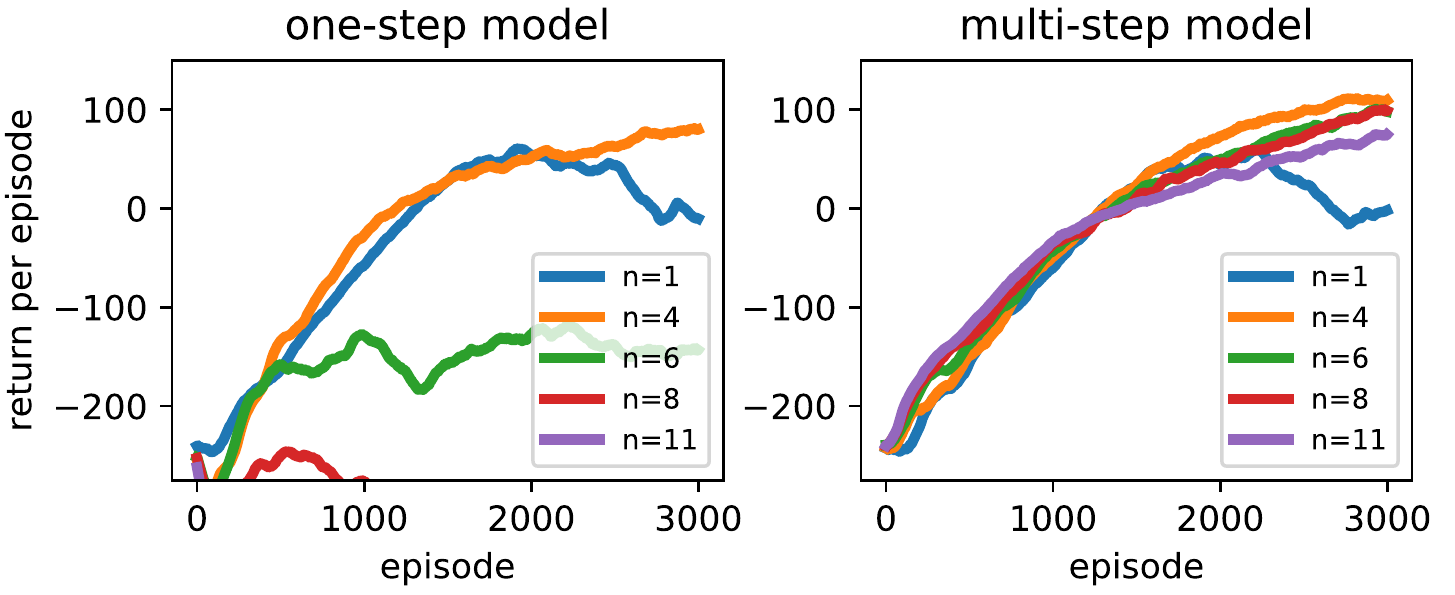}
\end{minipage}%
\begin{minipage}{.35\textwidth}
  \centering
  \includegraphics[width=.9\linewidth]{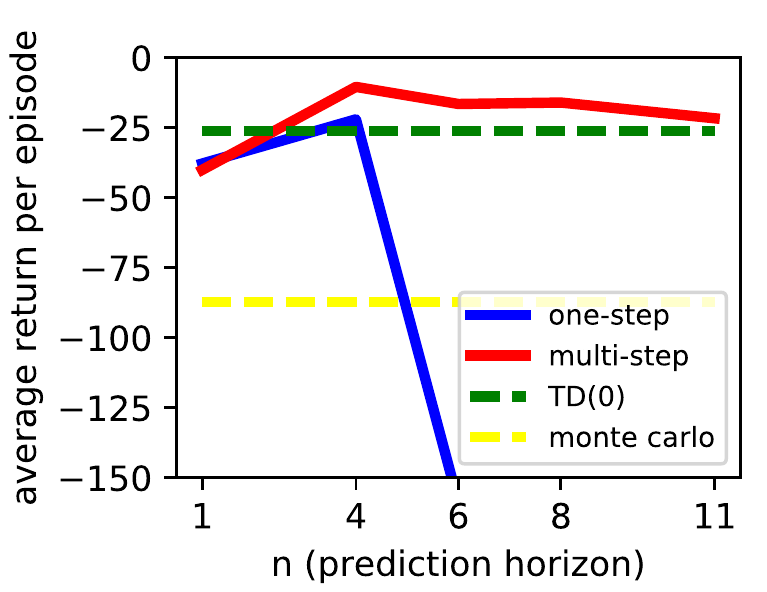}

\end{minipage}
  \caption{Performance of actor critic based on different ways of training the critic on Cart Pole (first row), Acrobot (second row), and Lunar Lander (third row). The first column shows performance with a one-step model, and the second column shows performance with a multi-step model-based. The third column shows area under the curve. Results are averaged over 50 runs. Higher is better.}
\label{model_based_rl_results}
\end{figure}

Although low squared error may be an indication of a good model, we actually care about effectiveness of a model during planning. To this end, we used the models for planning, specifically for value-function optimization. Recall that given a batch of experience $\mathcal{D}$ many reinforcement learning algorithms perform value-function optimization using some variant of the following update rule:

$$\theta \leftarrow \theta +\alpha\sum_{s,a\in \mathcal{D}} \big(G_{(s,a)}-\widehat Q(s,a;\theta)\big)^2 \nabla_{\theta}\widehat Q(s,a;\theta)\ .$$
For example, TD(0) uses the target $G_{(s,a)}=r+\gamma \widehat Q(s',a';\theta)$. Alternatively, using the multi-step model we can compute $G_{(s,a)}$ as shown in Algorithm \ref{alg:rollout_with_multi_step_model}. We can then compare the effectiveness of value-function optimization using the two models as well as using model-free baselines. Note that to ensure a fair comparison, we performed a fixed number of updates to the critic network, and just varied the type of updates. Results are presented in Figure \ref{model_based_rl_results}.
\begin{algorithm}
input a state $s_0$, an action $a_0$, a policy $\pi$, and a value-function estimate $\widehat{Q}$ \;
input a reward model $\widehat{R}$, and a multi-step transition model $\widehat{T}^{i}\ \forall i \in [1,n]$ \;
set hyper-parameters $n$ and $K$ \;
\SetAlgoLined
\For{$k=1:K$ }{
 \For{$i=0:n-1$}{
  $s_{i+1}\leftarrow \widehat{T}^{i+1}(s_{0},a_{0},...,a_{i}) $\;
  $r_{i}\leftarrow \widehat{R}(s_{i},a_{i},s_{i+1})$\;
  $a_{i+1}\sim \pi(\cdot|s_{i+1})$\Comment{sample an action from the policy}\;
 }
 $a_{n}\sim \pi(\cdot|s_{n})$\;
 $G_{k}\leftarrow (\sum_{i=0}^{n-1} \gamma^{i}r_{i})+\gamma^{n-1}\widehat{Q}(s_{n},a_{n})$  \Comment{estimated target from $k$th sample}\;
 }
  return $G_{(s_0,a_0)}=\frac{1}{K}\sum_{k=1}^{K}G_{k}$\Comment{return the average of $K$ samples}\;
 \caption{Computing the update target using a multi-step model.}
 \label{alg:rollout_with_multi_step_model}
\end{algorithm}

When planning with a one-step model, it was difficult to even outperform model-free learners, unless when we carefully chose the prediction horizon. On the other hand, planning with the multi-step model was clearly more effective than model-free baselines, and outperformed the one-step model except in the Acrobot domain and for small values of horizon. We also observed an inverted-U shape with respect to prediction horizon. We observe that there exists a trade-off: with short prediction horizon there is little value in look-aheads, and for very large prediction horiozn look-aheads can be misleading due to large error. Therefore, an intermediate value works best. \cite{jiang2015dependence} provided theoretical results that confirm this empirical observation.
\subsection{Atari Breakout}

Our desire is to use the multi-step model on larger domains. To this end, we trained one and multi-step models on Atari Breakout using its public implementation \citep{gym}. Our dataset consisted of episodes from a trained DQN with an epsilon-greedy policy. Our goal is to use the model for planning in future work, so we only focus on comparing multi-step prediction error here.

The architecture of our one-step and multi-step models were mainly based on \cite{oh2015action}. It consisted of an encoder and a decoder which predicts a frame n steps ahead. For the multi-step model, sequence of actions were fed into n distinct linear layers, then integrated into the network through a multiplication gate as they do in \cite{oh2015action}. We trained the one-step model and multi-step models, by minimizing mean squared error, on a dataset of 100 episodes of environmental interaction. We then tested the models on a held-out dataset. 
\begin{figure}
\centering
\begin{minipage}{.4\textwidth}
  \centering
  \includegraphics[width=1\linewidth]{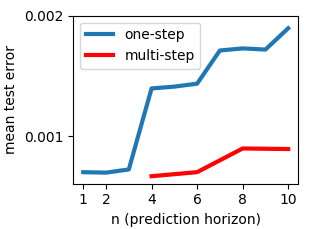}
  
\end{minipage}
  \caption{Accuracy of one-step and multi-step models on Atari Breakout.}
  \label{fig:atari}
\end{figure}

\begin{figure}
\centering
\begin{minipage}{.9\textwidth}
  \centering
  \includegraphics[width=.9\linewidth]{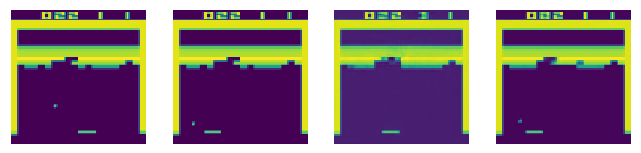}
\end{minipage}
\begin{minipage}{.9\textwidth}
  \centering
  \includegraphics[width=.9\linewidth]{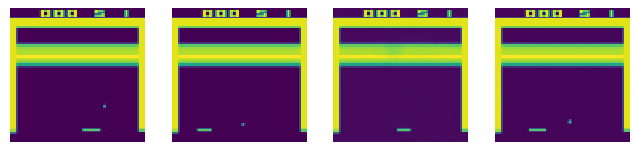}
\end{minipage}
\caption{Two examples of Atari Breakout results sampled from a held-out test set. On the first two column we see the input and (the true) output frames. The third column is a 10-step prediction made by the one-step model, and the last column is a 10-step prediction using the multi-step model. Clearly the multi-step model provides more useful predictions. We observed that the one-step model fails to accurately capture the ball in every example of our test set.}
\label{atari_examples}
\end{figure}

As shown in Figure \ref{fig:atari}, we observed that the one-step model performs well for the first 3 time steps, but then becomes inaccurate very quickly. Rollouts using the one-step model did not accurately capture ball and paddle location as seen in Figure \ref{atari_examples}. On the other hand, the multi-step model accurately predicted the location of the paddle (and sometimes the ball). The multi-step models' error is much lower than that of the one-step model, although the models increase in error slightly as the horizon increases. This result, though preliminary, is an evidence that the multi-step model can be more effective for planning in Atari.

\section{Related Work}
There has been numerous previous studies concerning approximate model-based reinforcement learning. Linear models were one of the earliest classes of approximate models studied in the literature \citep{parr_linear_models,linear_dyna}. The focus of these papers was learning a single-step linear model and the solution found when planning with the model. Their fundamental result was to show that the value function found by planner is the same as the value function found by linear temporal difference learning. Consistent with our findings, these papers reported little empirical benefits from one-step models.

\cite{Pilco} argued that naive planning without explicitly incorporating model uncertainty can be a problem. They proposed a model-based policy-search method based on gaussian processes to come up with a closed-form policy-gradient update. This line of research has recently become more popular \citep{gal_improving_pilco}, but note that in these studies planning was performed using a one-step model. We hypothesize that our multi-step model can also benefit from uncertainty encorporation during planning.

The compounding error problem has been observed in multiple prior studies \citep{time_series_error_compound,Talvitie14,Talvitie_self_correcting,Asadi_Lipschitz}. In the context of time series, an algorithm was presented that improves multi-step predictions by training the model on its own outputs \citep{time_series_error_compound}. \cite{Talvitie14} presented a similar algorithm for reinforcement learning, referred to it as hallucination, and showed that in some cases the new training scheme outperforms one-step methods without the hallucination technique. Later on this idea was theoretically investigated \citep{Talvitie_self_correcting}, and it was shown that pathological cases may arise if the policy is stochastic.

One can combine model-based and model-free approaches \citep{Dyna_original,yao_multi_dyna,combinations_kavosh,combinations_nagabandi}. The motivation here is to get the benefits of both approaches. Interestingly, it is recently discovered that the process of decision making in the brain is also implemented using two processes that closely resemble model-free (habitual) and model-based (goal-based) learning. Two existing theories are that the two processes compete \citep{daw_uncertainty} or cooperate \citep{gershman_retrospective} to make the final decisions.

Another interesting line of work is to learn models that, though imperfect, are tailored to the specific planning algorithm that is going to use them. For example, the model can be wrong, but it can still be very useful in terms of computing the fixed point of value function \citep{farahmand_value_aware,pires_error_bounds}. This idea is closely related to minimizing the Wasserstein metric as the loss function \citep{asadi_equivalence}. Two recent papers \citep{silver_predictron,oh_vpn} proposed a similar idea, namely to learn abstract state representations useful to predict values of states observed multiple timesteps ahead .

A few prior works have considered multi-step models. Perhaps the first attempt was presented by \cite{sutton_world_modeling} in which tabular multi-step models were studied. Later on \citep{options} proposed options, closed-loop policies with start and termination conditions. They then discussed the idea of learning option models in the tabular case. The idea of learning option models was later extended to the linear case \citep{liner_options}. The main limitation here is that we learn a model per option, and so when a new option is considered, a new model should be learned from scratch. Finally, \cite{vanseijen_deeper} articulated a multi-step model in the linear case and showed connections to temporal difference learning with eligibility traces. However the model is only valid for the current policy which is a limitation during planning.

In terms of empirical results, successful attempts to use one-step approximate models are rare. Indeed it is possible to learn reasonable one-step models of Atari games \citep{oh2015action}, as well as other challenging domains \citep{feinberg2018model}, but these models are usually not very useful for multi-step rollouts. \cite{machado2018revisiting} noted that the main issue is the compounding error problem associated with one-step models. This is consistent with our findings on the preliminary Atari experiment, and so, our multi-step model can provide a promising path to successful planning on Atari games by avoiding the compounding error problem.

We have noticed that the idea of learning multi-step models conditioned on action sequences is being explored in an independent and concurrent work\footnote{This work is an anonymous \href{https://openreview.net/pdf?id=B1g29oAqtm}{\underline{\textbf{submission}}} at ICLR 2019.}. The work focuses on a cross-entropy method that, inspired by model predictive control, learns best action-sequences by optimizing from a candidate list of action sequences provided to the agent. In contrast, we used the multi-step model for value-function optimization for actor-critic in which an explicit policy representation was used during planning. 
\section{Conclusion and Future Work}
We presented a simple approach to multi-step model-based reinforcement learning as an alternative for one-step model learning. We found the multi-step model to be more useful than the one-step model on domains considered in this paper. We believe that the discovery of this model is an important step towards model-based reinforcement learning in the approximate setting.

Consistent with previous work, we found that composition of imperfect one-step models can in general be catastrophic. Moreover, we found that multi-step rollouts are necessary in order to get the real benefits of planning. Based on our results, we believe that one-step models offer limited merits, and that composition of one-step models should in general be avoided.

The experiments reported in this work are still preliminary. That said, we believe that the reported results make a strong case for the multi-step model.

\newpage
\bibliographystyle{apalike}
\bibliography{nips_18_workshop}

\newpage
\section{Appendix}
Here we report hyperparameters to help readers reproduce our results. Upon publication of the paper, we will publicize our code as well.
\small
\begin{table}
\begin{tabular}{|c|c|c|c|}
\hline
                                                                                                          & Cart Pole                    & Acrobot                         & Lunar Lander           \\ \hline
$\gamma$ (discount rate)                                                                                            & 0.9999                       & 0.9999                          & 0.9999                 \\ \hline
critic network: hidden layers                                                                             & 1                            & 1                               & 1                      \\ \hline
\begin{tabular}[c]{@{}c@{}}critic network: \\ neurons per hidden layer\end{tabular}                       & 64                           & 64                              & 64                     \\ \hline
\begin{tabular}[c]{@{}c@{}}critic network: \\ step size (tuned from the list)\end{tabular}            & {[}0.1, 0.05, 0.025, 0.01{]} & {[}0.005,0.0025,0.001,0.0005{]} & {[}0.025,0.01,0.005{]} \\ \hline
\begin{tabular}[c]{@{}c@{}}critic network: \\ batches of update per episode\end{tabular}                  & 20                           & 20                              & 20                     \\ \hline
critic network: batch size                                                                                & 32                           & 32                              & 32                     \\ \hline
actor network: hidden layers                                                                              & 1                            & 1                               & 1                      \\ \hline
\begin{tabular}[c]{@{}c@{}}actor network: \\ neurons per hidden layer\end{tabular}                        & 64                           & 64                              & 64                     \\ \hline
actor network: step size                                                                                  & 0.005                        & 0.0005                          & 0.001                  \\ \hline
\begin{tabular}[c]{@{}c@{}}actor network: \\ batches of update per episode\end{tabular}                   & 1                            & 1                               & 1                      \\ \hline
actor network: batch size                                                                                 & length of episode            & length of episode               & length of episode      \\ \hline
transition functions: hidden layers                                                                       & 1                            & 1                               & 1                      \\ \hline
\begin{tabular}[c]{@{}c@{}}transition functions: \\ neurons per hidden layer\end{tabular}                 & 64                           & 64                              & 64                     \\ \hline
\begin{tabular}[c]{@{}c@{}}transition functions: step size\\  (optimized for one-step model)\end{tabular} & 0.001                        & 0.01                            & 0.001                  \\ \hline
\begin{tabular}[c]{@{}c@{}}transition functions: \\ batches of update per episode\end{tabular}            & 100                          & 20                              & 100                    \\ \hline
transition functions: batch size                                                                          & 128                          & 1024                            & 128                    \\ \hline
reward model: hidden layers                                                                               & 1                            & 1                               & 1                      \\ \hline
\begin{tabular}[c]{@{}c@{}}reward model: \\ neurons per hidden layer\end{tabular}                         & 128                          & 128                             & 128                    \\ \hline
reward model: step size                                                                                   & 0.1                          & 0.1                             & 0.00025                \\ \hline
\begin{tabular}[c]{@{}c@{}}reward model: \\ batches of update per episode\end{tabular}                    & 10                           & 10                              & 10                     \\ \hline
reward model: batch size                                                                                  & 128                          & 128                             & 128                    \\ \hline
maximum buffer size                                                                                       & 8000                         & 5000                             & 20000                  \\ \hline
target network update frequency                                                                           & 1 episode                    & 1 episode                       & 1 episode              \\ \hline
\end{tabular}
\caption{Hyper parameters used for control domains.}
\end{table}

\begin{table}
\centering
\begin{tabular}{|l|l|}
\hline
batch size                    & 4     \\ \hline
down sampling residual blocks & 4     \\ \hline
up sample residual blocks     & 4     \\ \hline
residual unit size            & 128   \\ \hline
latent size                   & 2048  \\ \hline
final conv size               & 32    \\ \hline
adam learning rate            & .0001 \\ \hline
\end{tabular}
\caption{Hyperparameters for Atari Experiments}
\end{table}
\normalsize
\end{document}